\documentclass[runningheads]{llncs}
\usepackage{graphicx}
\usepackage{amsmath}
\usepackage{amsfonts}
\usepackage{xcolor}
\usepackage{subcaption}
\usepackage{subfiles}
\usepackage{cleveref}
\usepackage[export]{adjustbox}
\usepackage{caption}
\usepackage{subcaption}

\usepackage{array}
\usepackage{amssymb}
\usepackage{float}
\begin{document}
\title{Multi-modal Variational Autoencoders for normative modelling across multiple imaging modalities}

\titlerunning{Multi-modal normative modelling}

\author{Ana Lawry Aguila
\and
James Chapman
\and
Andre Altmann
}

\authorrunning{A. Lawry Aguila et al.}

\institute{University College London, London, WC1E 6BT
\email{ana.aguila.18@ucl.ac.uk}\\ }

\maketitle              
\begin{abstract}
One of the challenges of studying common neurological disorders is disease heterogeneity including differences in causes, neuroimaging characteristics, comorbidities, or genetic variation. Normative modelling has become a popular method for studying such cohorts where the ‘normal’ behaviour of a physiological system is modelled and can be used at subject level to detect deviations relating to disease pathology. For many heterogeneous diseases, we expect to observe abnormalities across a range of neuroimaging and biological variables. However, thus far, normative models have largely been developed for studying a single imaging modality. We aim to develop a multi-modal normative modelling framework where abnormality is aggregated across variables of multiple modalities and is better able to detect deviations than uni-modal baselines. We propose two multi-modal VAE normative models to detect subject level deviations across T1 and DTI data. Our proposed models were better able to detect diseased individuals, capture disease severity, and correlate with patient cognition than baseline approaches. We also propose a multivariate latent deviation metric, measuring deviations from the joint latent space, which outperformed feature-based metrics.

\end{abstract}

\section{Introduction}\label{Introduction}
Normative modelling is a popular method to study heterogeneous brain disorders. Normative models assume disease cohorts sit at the tails of a healthy population distribution and quantify individual deviations from healthy brain patterns. Typically, a normative analysis constructs a normative model per variable, e.g., using Gaussian Process Regression (GPR)\cite{Marquand:2016}. Recently, to model complex non-linear interactions between features, deep-learning approaches using adversarial (AAE) and variational autoencoder (VAE) models have been proposed\cite{Pinaya:2021,LawryAguila:2022}. These models have a uni-modal structure with a single encoder and decoder network. So far, almost all deep-learning normative models have modelled only one modality. However, many brain disorders show deviations from the norm in features of multiple imaging modalities to a varying degree. Often it is unknown which modality will be the most sensitive. Thus, it is advantageous to develop normative models suitable for multiple modalities. 

Most previous deep-learning normative models measure deviations in the feature space\cite{Pinaya:2021,Kumar:2021}. However, for multi-modal models built from modalities containing highly different, but complementary information (e.g., T1 and DTI features as used here), we may not expect to see significantly greater deviations in the feature space compared to uni-modal methods. Indeed previous work has shown that, when using VAEs, even for one modality, measuring deviation in the latent space outperforms metrics in the feature space\cite{LawryAguila:2022} and provides a single measure of abnormality. As such, we develop a latent deviation metric suitable to measuring deviations in multi-modal data.

There are many approaches to extending VAEs to integrate information from multiple modalities and learn informative joint latent representations. Most multi-modal VAE frameworks learn separate encoder and decoder networks for each modality and aggregate the encoding distributions to learn a joint latent representation. Wu and Goodman\cite{Wu:2018} introduced a multi-modal VAE (mVAE) where each encoding distribution is treated as an ‘expert' and the Product-of-Experts (PoE), which takes a product of the experts' densities, is used to approximate a joint encoding distribution. The PoE approach treats all experts as equally credible taking a uniform contribution from every modality. In practice, however, different levels of noise, complexity and information are present in different modalities. Furthermore, the PoE joint distribution may be biased towards overconfident but miscalibrated experts leading to a sub-optimal joint representation. Shi et al.\cite{Shi:2019} address this problem by combining latent representations across modalities using a Mixture-of-Experts (MoE) approach and thus taking a vote amongst experts.  Alternatively, we propose a mVAE modelling the joint encoding distribution as a generalised Product-of-Experts (gPoE)\cite{Cao:2014}. We optimise modality specific weightings to account for different information content between experts and enable the model to ignore modalities, that are uninformative to a particular latent vector.

As far as we are aware, only one other multi-modal VAE normative modelling framework has been proposed in the literature which uses the PoE (PoE-normVAE)\cite{Kumar:2021}. However, Kumar et al.\cite{Kumar:2021} rely on measuring deviations in the feature space, which we argue does not leverage the benefits of multi-modal models. Here, we present an improved factorisation of the joint representation by modelling it as a weighted product or sum of each encoding distribution.  

Our contributions are two-fold. Firstly, we present two novel multi-modal normative modelling frameworks, MoE-normVAE and gPoE-normVAE, which capture the joint distribution between different imaging modalities. Our proposed models outperform baseline methods on two neuroimaging datasets. Secondly, we present a deviation metric, based on the latent space, suitable for detecting deviations in multi-modal normative distributions. We show that our metric better leverages the benefits of multi-modal normative models compared to feature space-based metrics.

\section{Methods}\label{Methods}
\subsubsection{Multi-modal Variational autoencoder (mVAE).}\label{mVAE}
Let $\textbf{X}=\{\textbf{x}_m\}^M_{m=1}$ be the observations of $M$ modalities. We use a mVAE to learn a multi-modal generative model (Figure \ref{fig:framework}c), where modalities are conditionally independent given a common latent variable, of the form $
p_\theta(\mathbf{X}, \mathbf{z})=p(\mathbf{z}) \prod_{m=1}^M p_{\theta_m}\left(\mathbf{x}_m \mid \mathbf{z}\right)
$. The likelihood distributions $p_{\theta_{m}}\left(\textbf{x}_{m} \mid \textbf{z}\right)$ are parameterised by decoder networks with parameters $\theta = \{ \theta_{1}, \ldots, \theta_{M} \}$. The goal of VAE training is to maximise the marginal likelihood of the data. However, as this is intractable, we instead optimise an evidence lower bound (ELBO):
\begin{equation}
\mathcal{L} = \mathbb{E}_{q_{\phi}(\textbf{z} \mid \textbf{X})}\left[\sum_{m=1}^{M} \log p_{\theta}\left(\textbf{x}_{m} \mid \textbf{z}\right)\right]-D_{K L}\left(q_{\phi}(\textbf{z} \mid \textbf{X})| p(\textbf{z})\right)
\end{equation}where the second term is the KL divergence between the approximate joint posterior $q_{\phi}\left(\textbf{z} \mid \textbf{X}\right)$ and the prior $p(\textbf{z})$. We model the posterior, likelihood, and prior distributions as isotropic gaussians. 

\subsubsection{Approximate joint posterior.} To train the mVAE, we must specify the form of the joint approximate posterior $q_{\phi}\left(\textbf{z} \mid \textbf{X}\right)$. Wu and Goodman\cite{Wu:2018} choose to factorise the joint posterior as a Product-of-Experts (PoE); $    q_{\phi}\left(\textbf{z} \mid \textbf{X}\right) = \frac{1}{K} \prod_{m=1}^{M} q_{\phi_{m}}\left(\textbf{z} \mid\textbf{x}_{m}\right)$, where the experts, i.e., individual posterior distributions $q_{\phi_{m}}\left(\textbf{z} \mid\textbf{x}_{m}\right)$, are parameterised by encoder networks with parameters $\phi = \{ \phi_{1}, \ldots, \phi_{M} \}$. $K$ is a normalisation term. Assuming each encoder network follows a Gaussian distribution $q\left(\textbf{z} \mid \textbf{x}_{m}\right)=\mathcal{N}(\boldsymbol{\mu}_m, \boldsymbol{\sigma}_{m}^{2} \textbf{I})$, the parameters of joint posterior distribution can be computed\cite{Hwang:2021}; $
\boldsymbol{\mu} = \frac{\sum_{m=1}^{M} \boldsymbol{\mu}_{m} / \boldsymbol{\sigma}_{m}^{2}}{\sum_{m=1}^{M} 1 / \boldsymbol{\sigma}_{m}^{2}} \quad \text { and } \quad \boldsymbol{\sigma}^{2} =\frac{1}{\sum_{m=1}^{M} 1 / \boldsymbol{\sigma}_{m}^{2}}
$ (see Supp. for proofs).

However, overconfident but miscalibrated experts may bias the joint posterior distribution (see Figure \ref{fig:framework}b) which is undesirable for learning informative latent representations between modalities\cite{Shi:2019}. 

Shi et al.\cite{Shi:2019} instead factorise the approximate joint posterior as a Mixture-of-Experts (MoE); $q_{\Phi}\left(\textbf{z} \mid \textbf{X}\right)= \frac{1}{K} \sum_{m=1}^{M} \frac{1}{M} q_{\phi_{m}}\left(\textbf{z} \mid \textbf{x}_{m}\right).$

In the MoE setting, each uni-modal posterior $q_{\phi}(\textbf{z} \mid \textbf{x}_m)$ is evaluated with the generative model $p_{\theta}\left(\textbf{X}, \textbf{z}\right)$ such that the ELBO becomes:
\begin{equation}
\mathcal{L} = \sum_{m=1}^{M}\left[\mathbb{E}_{q_{\phi}(\textbf{z} \mid \textbf{x}_{m})}\left[\sum_{m=1}^{M} \log p_{\theta}\left(\textbf{x}_{m} \mid \textbf{z}\right)\right]-D_{K L}\left(q_{\phi}(\textbf{z} \mid \textbf{x}_m)| p(\textbf{z})\right)\right].
\end{equation}
But this approach only takes each uni-modal encoding distribution separately into account during training. Thus there is no explicit aggregation of information from multiple modalities in the latent representation for reconstruction by the decoder networks. For modalities with a high degree of modality-specific variation, this enforces an undesirable upperbound on the ELBO potentially leading to a sub-optimal approximation of the joint distribution\cite{Daunhawer:2021}.

\subsubsection{Generalised Product-of-Experts joint posterior.} We propose an alternative approach to mitigate the problem of overconfident experts by factorising the joint posterior as a generalised Product-of-Experts (gPoE)\cite{Cao:2014}; $q_{\phi}\left(\textbf{z} \mid \textbf{X}\right) = \frac{1}{K} \prod_{m=1}^{M} q_{\phi_{m}}^{\alpha_{m}}\left(\textbf{z} \mid\textbf{x}_{m}\right)$, where $\alpha_{m}$ is a weighting for modality $m$ such that $\sum_{m=1}^{M}\alpha_{m} = 1$ for each latent dimension and $0<\alpha_{m}<1$. We optimise $\alpha$ during training allowing the model to weight experts in such a way as to learn an approximate joint posterior $q_{\phi}\left(\textbf{z} \mid \textbf{X}\right)$ where the likelihood distribution $p_{\theta}\left(\textbf{X} \mid \textbf{z}\right)$ is maximised. This provides a means to down-weigh overconfident experts. Furthermore, as $\alpha$ is learnt per latent dimension, different modality weightings can be learnt for different vectors, thus explicitly incorporating modality specific variation in addition to shared information in different dimensions of the joint latent space. Similarly to the PoE approach, we can compute the parameters of the joint posterior distribution; $
\boldsymbol{\mu} = \frac{\sum_{m=1}^{M} \boldsymbol{\mu}_{m}\boldsymbol{\alpha}_{m} / \boldsymbol{\sigma}_{m}^{2}}{\sum_{m=1}^{M} \boldsymbol{\alpha}_{m} / \boldsymbol{\sigma}_{m}^{2}} \quad \text { and } \quad \boldsymbol{\sigma}^{2} = \sum_{m=1}^{M} \frac{1}{ \boldsymbol{\alpha}_{m} / \boldsymbol{\sigma}_{m}^{2}}
$.

Recently, a gPoE mVAE was proposed for learning joint representations of hand-poses and surgical videos\cite{Joshi:2022}. However, we emphasize that our approach differs in application and offers a more lightweight implementation (Joshi et al.\cite{Joshi:2022} require training of auxiliary networks to learn $\alpha$ per sample).

\subsubsection{Multi-modal normative modelling.} 
\begin{figure}
     \centering
     \begin{minipage}[t]{.82\linewidth}
         \centering
         \subcaptionbox{}{\includegraphics[width=\textwidth,trim={1cm 0.5cm 0 1cm}, scale=0.8]{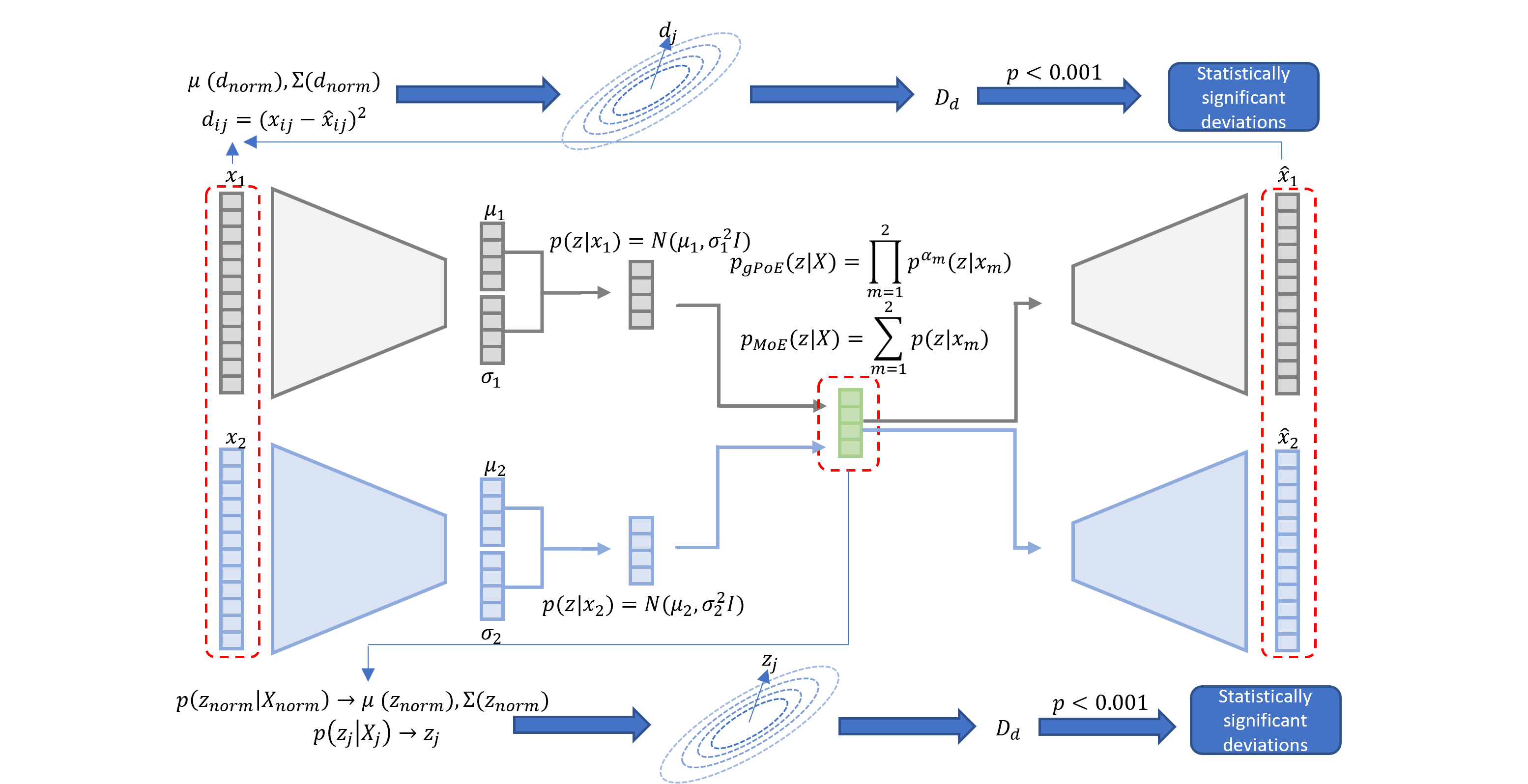}\label{fig:normative_framework}}
        
     \end{minipage}
     \begin{minipage}[b]{.143\linewidth}
         \centering
            \subcaptionbox{}{\includegraphics[width=\textwidth,trim={2.5cm 3cm 0cm 5cm}, scale=0.005]{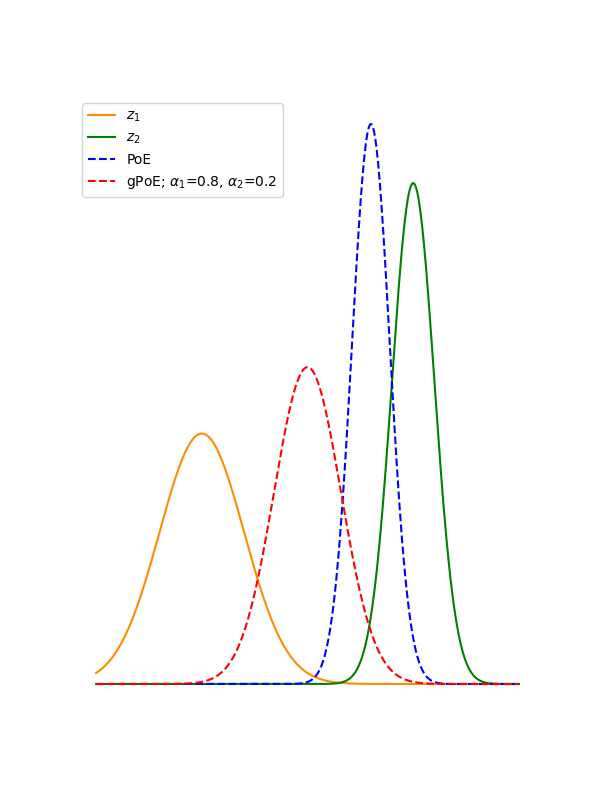}\label{fig:joint_dist}}
         \centering
            \subcaptionbox{}{\includegraphics[width=\textwidth,trim={1cm 3cm 0cm 0}, scale=0.005]{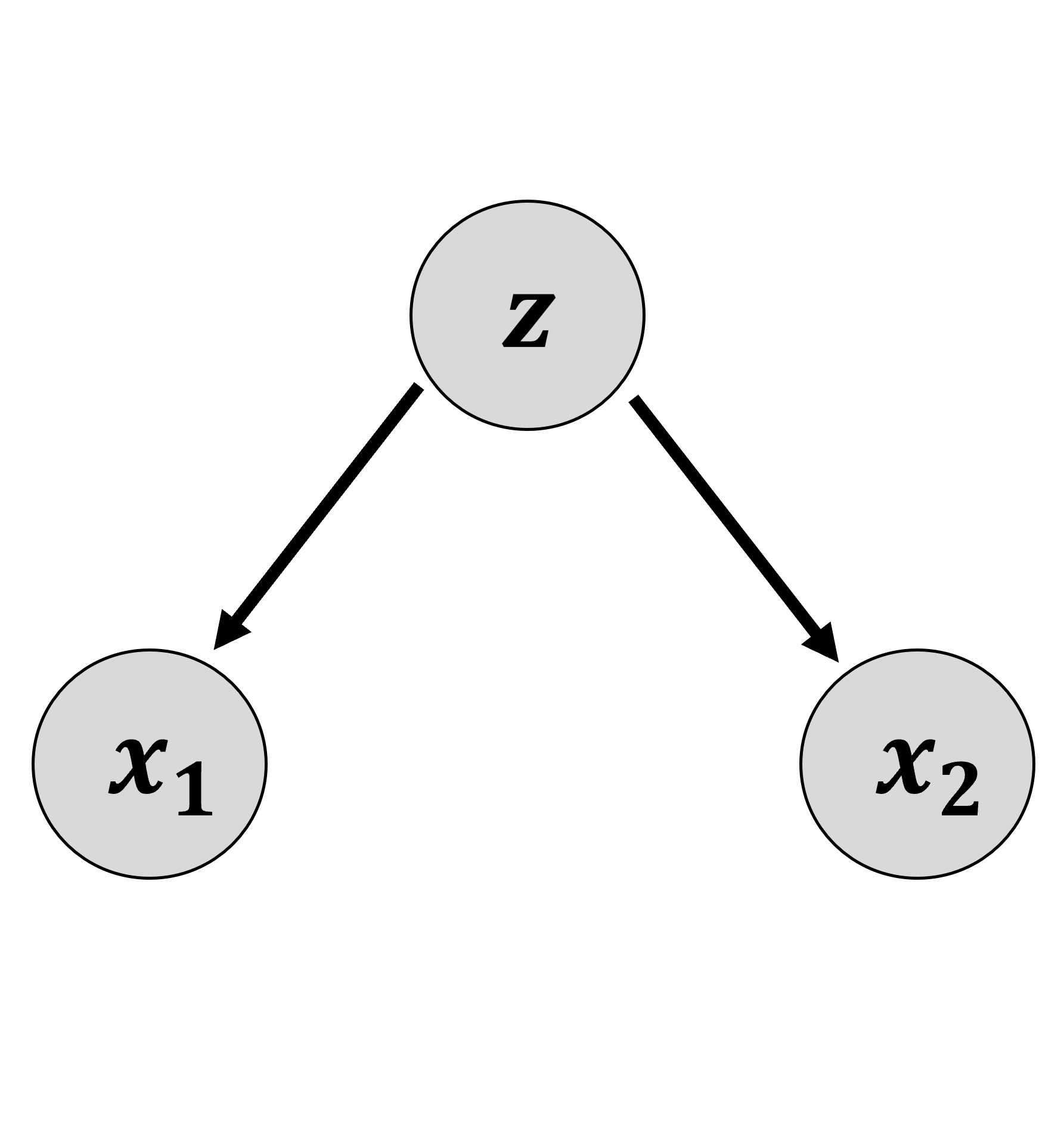}\label{fig:generative_model}}
     \end{minipage}
        \caption{(a) gPoE-normVAE and MoE-normVAE normative framework. All normVAE models were implemented using parameter settings; maximum epochs=2000, batch size=256, learning rate=$10^{-4}$, early stopping=50 epochs, encoder layers=[20, 40], decoder layers=[20, 40]. A  ReLU activation function was applied between layers. Models were trained with a range of latent space sizes ($L_{\text{dim}}$) from 5 to 20. Models with $L_{\text{dim}}$=10 were fine-tuned (maximum 100 epochs) using the ADNI healthy cohort. Learnt $\alpha$ values are given in Supp. Table 1. (b) Example PoE and gPoE joint distributions. (c) Graphical model.}\label{fig:framework}
\end{figure}

We propose two mVAE normative modelling frameworks shown in Figure \ref{fig:framework}a. MoE-normVAE, which uses a MoE joint posterior distribution, and gPoE-normVAE, which uses a gPoE joint posterior distribution. For both models, the encoder $\phi$ and decoder $\theta$ parameters are trained to characterise a healthy population cohort. normVAE models assume abnormality due to disease effects can be quantified by measuring deviations in the latent space\cite{LawryAguila:2022} or the feature space\cite{Pinaya:2021}. At test time, the clinical cohort is passed through the encoder and decoder networks. Deviations of test subjects from the multi-modal latent space of the healthy controls and data reconstruction errors are measured. We compare our methods to the previously proposed PoE-normVAE\cite{Kumar:2021} and three uni-modal models; two single modality and one multi-modality with a concatenated input.

To compare our normVAE models to a classical normative approach, we trained one GPR (using the PCNToolkit) per feature on a sub-set of 2000 healthy UK Biobank individuals and used extreme value statistics to calculate subject-level abnormality index\cite{Marquand:2016}.  We used a top 5\% abnormality threshold (set using the healthy training cohort) to calculate a significance ratio (see Equation \ref{sig_ratio}).

\subsubsection{Multi-modal latent deviation metric.}\label{sec:multi_modal_deviation_metric} Previous works using autoencoders as normative models mostly relied on feature-space based deviation methods\cite{Pinaya:2021,Kumar:2021}. That is, they compare the input value for subject $j$ for the $i$-th brain region $x_{ij}$ to the value reconstructed by the autoencoder $\widehat{x}_{ij}$: $d_{ij}=\left(x_{ij}-\widehat{x}_{ij}\right)^{2}$. Kumar et al.\cite{Kumar:2021} propose the following normalised z-score metric on the data reconstruction (a univariate feature space metric):
\begin{equation}
    D_{\text{uf}}=\frac{d_{i j}-\mu_{\text {norm}}\left(d_{i j}^{\text {norm}}\right)}{\sigma_{\text {norm}}\left(d_{i j}^{\text {norm}}\right)}
\end{equation}
where $\mu_{\text {norm}}\left(d_{i j}^{\text {norm}}\right)$ is the mean and $\sigma_{\text {norm}}\left(d_{i j}^{\text {norm}}\right)$ the standard deviation of the deviations $d_{i j}^{\text {norm}}$ of a holdout healthy control cohort.

However, in the multi-modal setting, feature space-based deviation metrics may not highlight the benefits of multi-modal models over their uni-modal counterparts. The goal of the joint latent representation is to capture information from all modalities. Thus, decoders for each modality must extract the information from the joint latent representation, which now carries information from all other modalities as well. Therefore, data reconstructions capture only information relevant to a particular modality and may also be poorer compared to uni-modal methods. As such, particularly when incorporating modalities with a high degree of modality-specific variation, we believe latent space deviation metrics would better capture deviations from normative behaviour across multiple modalities. Then, once an abnormal subject has been identified, feature space metrics can be used to identify deviating brain regions (e.g. Supp. Figure 3).  

We propose a latent deviation metric to measure deviations from the joint normative distribution. To account for correlation between latent vectors and derive a single multivariate measure of deviation, we measure the Mahalanobis distance from the encoding distribution of the training cohort:

\begin{equation}\label{latentdev}
D_{\text{ml}}=\sqrt{\left(z_{j}-\mu(z^{\text{norm}})\right)^T \Sigma(z^{\text{norm}})^{-1}\left(z_{j}-\mu(z^{\text{norm}})\right)} 
\end{equation}

where $z_j \sim q\left(\textbf{z}_j \mid \textbf{X}_j\right)$ is a sample from the joint posterior distribution for subject $j$, $\mu(z^{\text{norm}})$ is the mean and $\Sigma(z^{\text{norm}})$ the covariance of the healthy cohort latent position. We use robust estimates of the mean and covariance to account for outliers within the healthy control cohort. 
For closer comparison with $D_{\text{ml}}$, we derive the following multivariate feature space metric: 

\begin{equation}\label{datadev}
D_{\text{mf}}=\sqrt{\left(d_{j}-\mu(d^{\text{norm}})\right)^T \Sigma(d^{\text{norm}})^{-1}\left(d_{j}-\mu(d^{\text{norm}})\right)} 
\end{equation}
where $d_j = \{ d_{ij}, \ldots, d_{Ij} \}$ is the reconstruction error for subject $j$ for brain regions $(i=1,...,I)$, $\mu(d^{\text{norm}})$ is the mean and $\Sigma(d^{\text{norm}})$ the covariance of the healthy cohort reconstruction error. 
\subsubsection{Assessing deviation metric performance.} For each model, we calculated $D_{\text{ml}}$ and $D_{\text{mf}}$ for a healthy holdout cohort and disease cohort. For each deviation metric, we identified individuals whose deviations were significantly different from the healthy training distribution ($p<0.001$)\cite{Tabachnick:2018}. Ideally, we want a model which correctly identifies disease individuals as outliers and healthy individuals as sitting within the normative distribution. As such, we use the following significance ratio (positive likelihood ratio) to assess model performance:
\begin{equation}\label{sig_ratio}
\text{significance ratio} = \frac{\text{True positive rate}}{\text{False positive rate}} = \frac{\text{TPR}}{\text{FPR}}
=\frac{\frac{N_\text{disease}(\text{outliers})}{N_\text{disease}}}{\frac{N_\text{holdout}(\text{outliers})}{N_\text{holdout}}}
\end{equation}
In order to calculate significance ratios, we calculated $D_{\text{uf}}$ relative to the training cohort for the healthy holdout and disease cohorts (Bonferroni adjusted p=0.05/$N_{\text{features}}$)\cite{Kumar:2021}.

\section{Experiments}\label{Experiments}
\subsubsection{Data processing.}\label{Data processing}
To train the normVAE models, we used 10,276 healthy subjects from the UK Biobank\cite{Sudlow:2015}. We used pre-processed (provided by the UK Biobank\cite{Alfaro:2018}) grey-matter volumes for 66 cortical (Desikan-Killiany atlas) and 16 subcortical brain regions, and Fractional Anisotropy (FA) and Mean Diffusivity (MD) measurements for 35 white matter tracts (John Hopkins University atlas). At test time, we used 2,568 healthy controls from a holdout cohort and a cohort of 122 individuals with one of several neurodegenerative disorders; motor neuron disease, multiple sclerosis, Parkinson's disease, dementia/Alzheimer/cognitive-impairment and other demyelinating disease.

We also tested the models using an external dataset. We extracted 213 subjects from the Alzheimer's Disease Neuroimaging Initiative (ADNI)\cite{Petersen:2010} dataset with significant memory concern (SMC; N=27), early mild cognitive impairment (EMCI; N=63), late mild cognitive impairment (LMCI; N=34), Alzheimer's disease (AD; N=43) as well as healthy controls (HC; N=45). We used the healthy controls to fine-tune the models in a transfer learning approach. The same T1 and DTI features as for the UK Biobank were extracted for the ADNI dataset. 

Non-linear age and linear ICV affects where removed from the DTI and T1 MRI features of both datasets\cite{Pomponio:2020}. Each brain ROI was normalised by removing the mean and dividing by the standard deviation of the healthy control cohort brain regions.

\subsubsection{UK Biobank results.} As expected, we see greater significance ratios for all models when using $D_{\text{ml}}$ rather than $D_{\text{mf}}$ (Table \ref{fig:deviations_UKBB}). When using $D_{\text{mf}}$ or $D_{\text{uf}}$, all models perform similiarly. Using $D_{\text{ml}}$ over $D_{\text{mf}}$ leads to a 4-fold increase in the signficance ratio. Further, our proposed models give the best overall performance across different $L_{\text{dim}}$ with the highest significance ratio for gPoE-normVAE with $L_{\text{dim}}$=10. Generally, all multi-modal normVAE showed better performance than the uni-modal models suggesting that by modelling the joint distribution between modalities, we can learn better normative models.
\begin{table}[!htb]
    \caption{Significance ratio calculated from $D_{\text{ml}}$, $D_{\text{mf}}$, and $D_{\text{uf}}$ for the UK Biobank. See Supp. for results in figure form. Using GPR, we observed a significance ratio of 6.01, poorer performance than our models (using $D_{\text{ml}}$).}\label{fig:deviations_UKBB}
    \vspace*{0.2cm}
    \begin{subtable}{\linewidth}
      \centering
        \resizebox{\textwidth}{!}{%
\begin{tabular}{l@{\hspace{1.5\tabcolsep}}|llll||llll|llll}
 & \multicolumn{4}{l}{Significance ratio, $D_{\text{ml}}$} & \multicolumn{4}{l}{Significance ratio, $D_{\text{mf}}$} & \multicolumn{4}{l}{Significance ratio, $D_{\text{uf}}$} \\   

Latent dimension &                             5 &        10 &        15 &        20 &                             5 &        10 &        15 &        20 &              5 &        10 &        15 &        20 \\
\hline
gPoE-normVAE (ours)               &                          \textbf{7.89} &  \textbf{9.24} &  7.02 &   7.6 &                          1.62 &  1.62 &  1.61 &   1.7 &                          1.37 &  1.41 &  1.37 &  1.56 \\
MoE-normVAE (ours)                &                          7.25 &  8.77 &  \textbf{7.09} &  \textbf{7.94} &                          1.67 &   1.7 &  1.59 &  1.68 &                          1.44 &  1.46 &  1.46 &  1.44 \\
PoE-normVAE          &                           7.4 &  8.06 &  5.71 &  6.96 &                           1.6 &  1.62 &  1.66 &  1.71 &                          1.45 &  1.42 &  1.39 &  1.41 \\
concatenated normVAE   &                          7.63 &  6.21 &  5.43 &  3.55 &                          1.61 &  1.59 &  1.57 &  1.58 &                          1.48 &   1.4 &  1.39 &  1.44 \\
T1 normVAE             &                          5.26 &  4.43 &  2.63 &  2.44 &                               &       &       &       &                             
  &       &       &       \\
DTI normVAE           &                          6.82 &  7.35 &  3.96 &  2.61 &                               &       &       &       &                             
  &       &       &       \\
Average T1\&DTI normVAE &                               &       &       &       &                          1.63 &   1.7 &  1.64 &  1.64 &                          1.47 &  1.53 &  1.45 &  1.45 \\
\end{tabular}}
    \end{subtable}
\end{table}
\subsubsection{ADNI results.} 
\begin{figure}
     \centering
     \begin{subfigure}[b]{0.95\textwidth}
         \centering
          \includegraphics[width=\textwidth,trim={0 1.2cm 0 1.2cm}, scale=0.8]{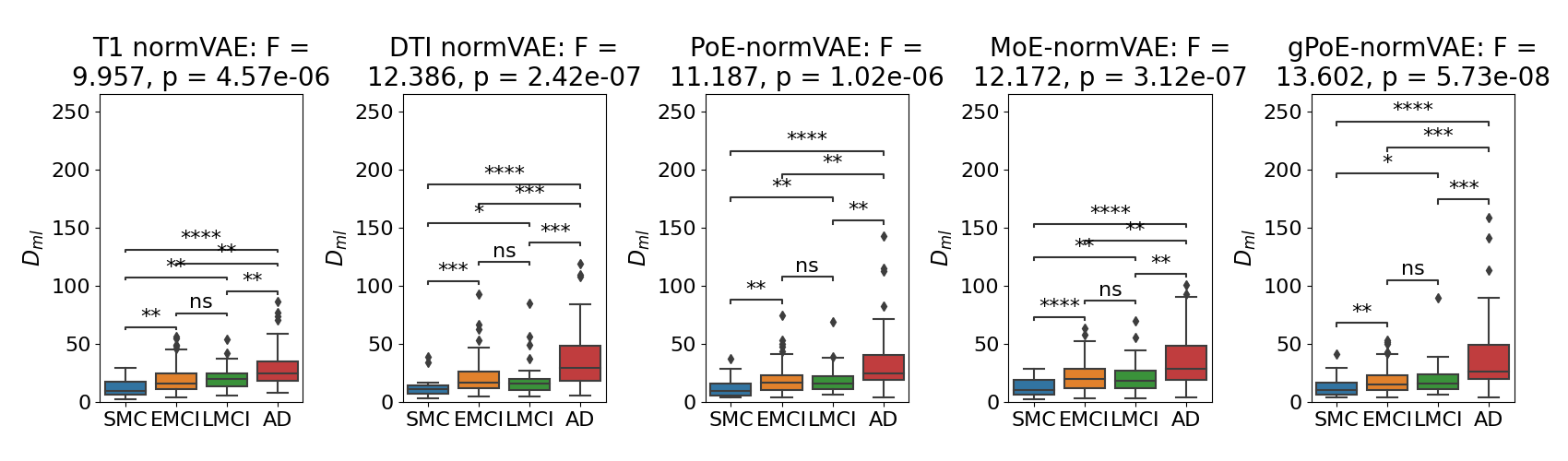}
          \caption{}
          \label{fig:ADNI_deviations_Dl}
     \end{subfigure}
     \adjustbox{valign=b}{\begin{subfigure}[b]{\textwidth}
    \begin{subtable}[b]{\linewidth}
      \centering
        \resizebox{\textwidth}{!}{%
\begin{tabular}{lrrrr}
       model &  Memory score (r) &  Memory score (p) &  Executive function score (r) &  Executive function score (p) \\
  T1 normVAE &         -0.201754 &          8.9334e-03 &                     -0.289296 &                  1.495292e-04 \\
 DTI normVAE &         -0.320889 &          2.3577e-05 &                     -0.406776 &                  4.898105e-08 \\
 PoE-normVAE &         -0.255954 &          8.4200e-04 &                     -0.404015 &                  6.148406e-08 \\
 MoE-normVAE &         -0.301793 &       7.3842e-05 &                     -0.396141 &                   1.1629e-07 \\
gPoE-normVAE &         \textbf{-0.335123} &          \textbf{9.5594e-06} &                     \textbf{-0.468410} &                  \textbf{1.730127e-10} \\
\end{tabular}}
    \end{subtable}
    \caption{}\label{fig:ADNI_pearsoncorr}
    \end{subfigure}}
        \caption{(a) $D_{\text{ml}}$ by disease label for the T1 normVAE, DTI normVAE, PoE-normVAE and gPoE-normVAE models ($L_{\text{dim}}$=10). Statistical annotations on the $D_{\text{ml}}$ values were generated using Welch's t-tests between pairs of disease groups;  $\text{ns}: 0.05< p <= 1, \ast: 0.01< p <=0.05, \ast\ast: 0.001< p <= 0.01, \ast\ast\ast: 0.0001< p <= 0.001, \ast\ast\ast\ast:p <= 0.0001$. Robust estimates of the mean and covariance were not used to calculate $D_{\text{ml}}$ due to the small healthy cohort size. (b) Pearson correlation between $D_{\text{ml}}$ and patient cognition represented by age adjusted memory and executive function composite scores.}\label{fig:deviations_ADNI}
\end{figure}

Previous work\cite{LawryAguila:2022} explored the ability of a uni-modal T1 normVAE to detect deviations in the ADNI cohorts. Figure \ref{fig:ADNI_deviations_Dl} shows the latent deviation $D_{\text{ml}}$ for different diagnosis in the ADNI cohort for the T1 normVAE, DTI normVAE, PoE-normVAE, MoE-normVAE and gPoE-normVAE models. All models reflect the increasing disease severity with increasing disease stage. The gPoE-normVAE model showed greater sensitivity to disease stage as suggested by the higher F statistic and p-values from an ANOVA analysis. We measured the Pearson correlation with composite measures of memory and executive function (Figure \ref{fig:ADNI_pearsoncorr}) and found that our proposed model exhibited greater correlation with both cognition scores than baseline approaches. Finally, we see that the sensitivity to disease severity for the gPoE-normVAE model extends to the feature space where we see a general increase in average $D_{\text{uf}}$ from the LMCI to AD cohort (Supp. Figures 3a and 3b respectively).
\section{Discussion and further work}\label{Discussion}
We have built on recent works\cite{Pinaya:2021,Kumar:2021,LawryAguila:2022} and introduced two novel mVAE normative models, which provide an alternative method of learning the joint normative distribution between modalities to address the limitations of current approaches. Our models provide a more informative joint representation compared to baseline methods as evidenced by the better significance ratio for the UK Biobank dataset and greater sensitivity to disease staging and correlation with cognitive measures in the ADNI dataset.  We also proposed a latent deviation metric suitable for detecting deviations in the multivariate latent space of multi-modal normative models which gave an approximately 4-fold performance increase over metrics based on the feature space. 

Further work will involve extending our models to more data modalities, such as genetic variants, to better characterise the behaviour of a physiological system. We note that, for fair comparison across models, we remove the effects of confounding variables prior to analysis. However, confounding effects could be removed during analysis via condition variables as done in previous work\cite{LawryAguila:2022}. 

Normative models have been successfully applied to the study of a range of heterogeneous diseases. Diseases often present abnormalities across a range of neuroimaging, biological and physiological features which provide different information about the underlying disease process. Normative systems that incorporate features from different data modalities offer a holistic picture of the disease and will be capable of detecting abnormalities across a broad range of different diseases. Furthermore, multi-modal normative modelling captures the relationship between different modalities in healthy individuals, with disruption to this relationship potentially leading to a disease signal. 
\subsubsection{Acknowledgements.}This work is supported by the EPSRC-funded UCL Centre for Doctoral Training in Intelligent, Integrated Imaging in Healthcare (i4health) and the Department of Health’s NIHR-funded Biomedical Research Centre at University College London Hospitals. 

Data used in preparation of this article were obtained from the Alzheimer’s Disease Neuroimaging Initiative (ADNI)  database  (adni.loni.usc.edu).  As  such,  the  investigators  within  the  ADNI  contributed  to  the  design and implementation of ADNI and/or provided data but did not participate in analysis or writing of this report. A complete listing of ADNI investigators can be found at:\url{http://adni.loni.usc.edu/wp-content/uploads/how\_to\_apply/ADNI\_Acknowledgement\_List.pdf}

%
%
%
\newpage
\bibliographystyle{splncs04}
\bibliography{main}

\newpage

\section*{\centering \Large Supplementary Materials}
\begin{table}[!htb]
    \caption{$\alpha$ weightings for the gPoE-normVAE model with $L_{\text{dim}}$=10.}\label{alpha_table}
    \vspace*{0.3cm}
    \begin{subtable}{\linewidth}
      \centering
        \resizebox{\textwidth}{!}{%
        \begin{tabular}{|l|r|r|r|r|r|r|r|r|r|r|}
        \hline
        modality &  latent 0 &  latent 1 &  latent 2 &  latent 3 &  latent 4 &  latent 5 &  latent 6 &  latent 7 &  latent 8 &  latent 9 \\
        \hline
              T1 &  0.422716 &   0.319826 &  0.704999 &    0.4191 &  0.345487 &  0.520485 &  0.464519 &  0.698466 &  0.156988 &  0.632687 \\
             DTI &  0.577284 &   0.680174 &  0.295001 &    0.5809 &  0.654513 &  0.479515 &  0.535481 &  0.301534 &  0.843012 &  0.367313 \\
        \hline
        \end{tabular}}
    \end{subtable}
\end{table}
\begin{figure}[H]
     \centering
     \begin{subfigure}[b]{\textwidth}
         \centering
          \includegraphics[width=\textwidth,trim={0 0.5cm 0 4cm}, scale=1]{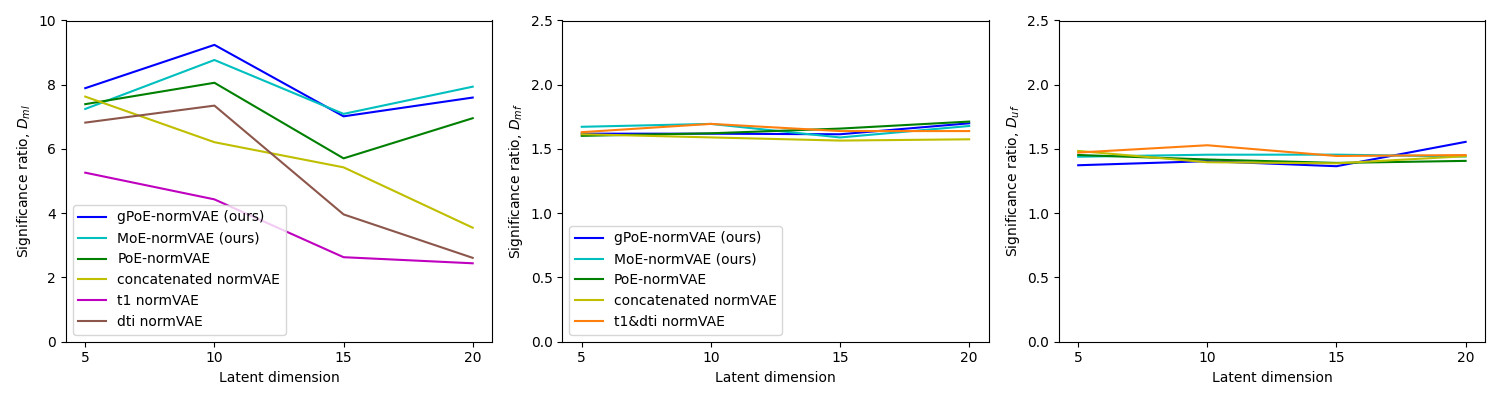}
          \caption{}
          \label{fig:deviations_Dl_Dd}
     \end{subfigure}
        \caption{(a) Significance ratio calculated from $D_{\text{ml}}$, and $D_{\text{mf}}$ and $D_{\text{uf}}$ for the UK Biobank.}\label{fig:deviations_UKBB}
\end{figure}
\begin{figure}
     \centering
     \begin{subfigure}[b]{\textwidth}
         \centering
          \includegraphics[width=\textwidth,trim={0 1cm 0 5cm}]{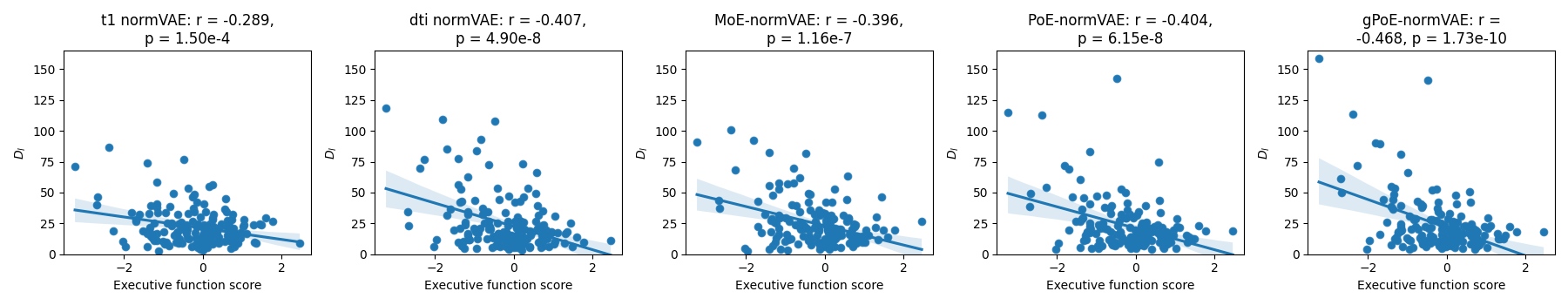}
          \caption{}
          \label{fig:ADNI_corr_EF}
     \end{subfigure}
     \begin{subfigure}[b]{\textwidth}
         \centering
          \includegraphics[width=\textwidth,trim={0 1cm 0 0cm}]{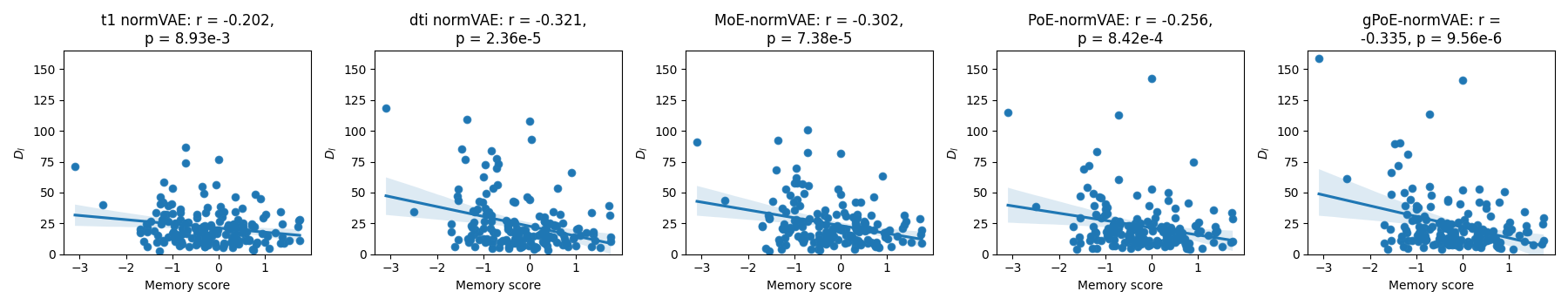}
          \caption{}
          \label{fig:ADNI_corr_MEM}
     \end{subfigure}
     \caption{Pearson correlation between $D_{\text{ml}}$ and (a) executive function and (b) memory scores for the T1 normVAE, PoE normVAE and gPoE normVAE models applied to the ADNI dataset.}
\end{figure}
\subsubsection{Product and generalised product of Gaussian's.}
In this section, we provide the derivation for the parameters of the Product of Experts (PoE) and generalised Product of Experts (gPoE) Gaussian product distributions.

\begin{figure}
     \centering
     \begin{subfigure}[b]{0.9\textwidth}
        \begin{minipage}{.5\textwidth}
          \centering
          \includegraphics[width=\textwidth,trim={0 2cm 1cm 3cm}]{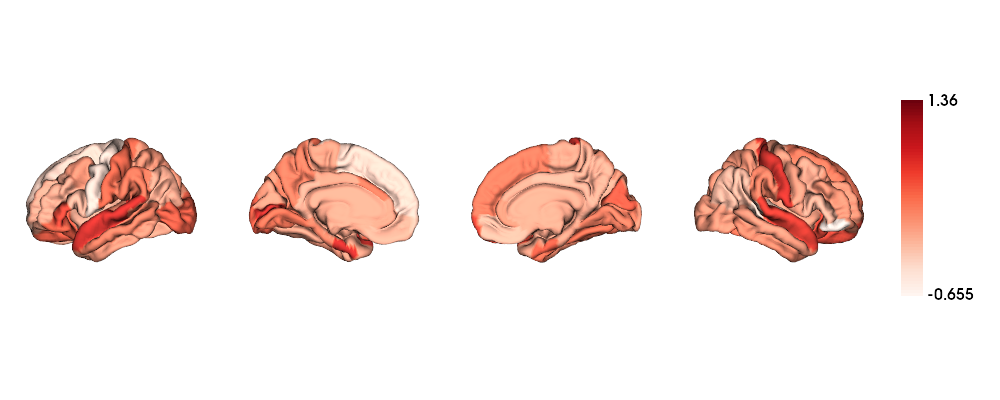}
            \includegraphics[width=\textwidth,trim={0 4cm 1cm 3cm}]{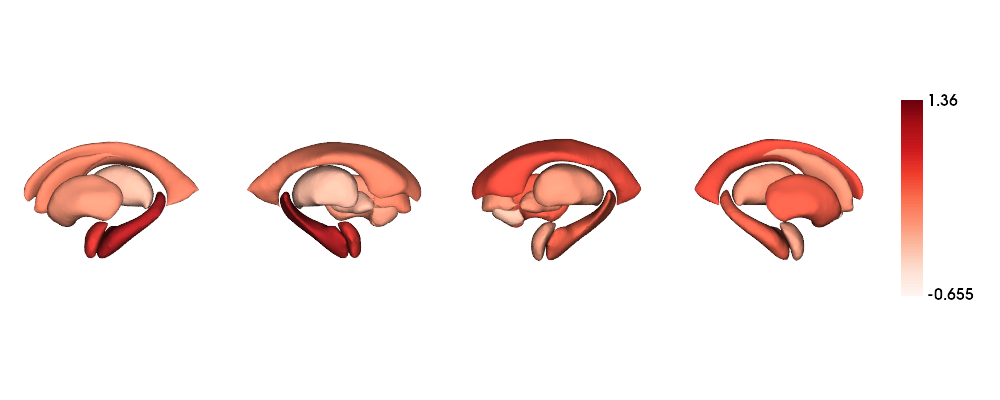}
        \end{minipage}%
        \begin{minipage}{.5\textwidth}
          \centering
          \includegraphics[width=0.49\textwidth,trim={0 4cm 3cm 3cm}]{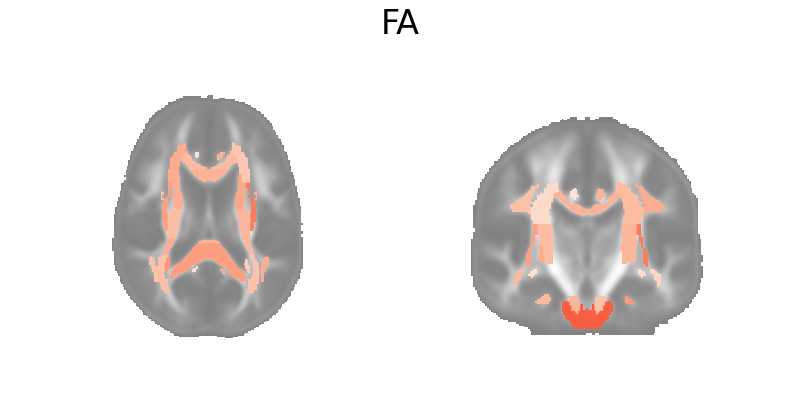}
            \includegraphics[width=0.49\textwidth,trim={0 4cm 3cm 3cm}]{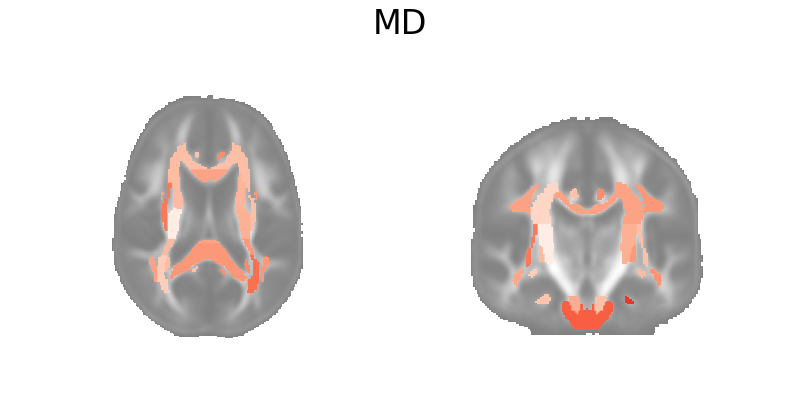}
        \end{minipage}
            \caption{}
          \label{fig:ADNI_LMCI_recon}
     \end{subfigure}
     \begin{subfigure}[b]{0.9\textwidth}
        \begin{minipage}{.5\textwidth}
          \centering
          \includegraphics[width=\textwidth,trim={0 2cm 1cm 3cm}]{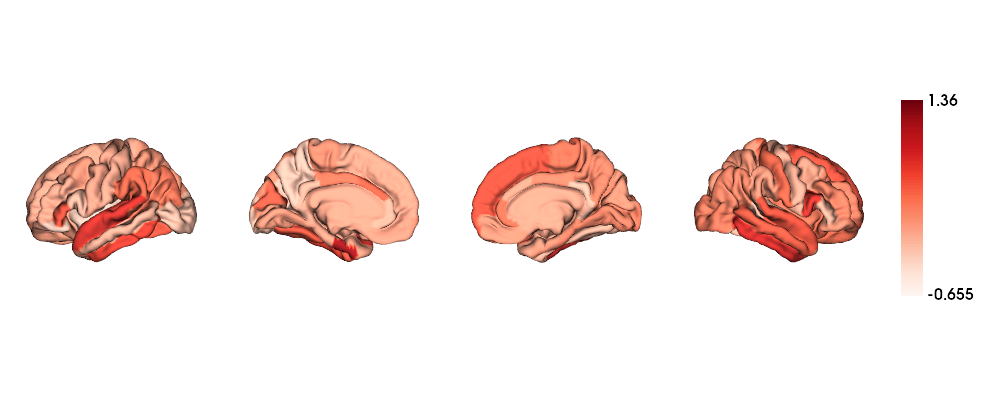}
            \includegraphics[width=\textwidth,trim={0 4cm 1cm 3cm}]{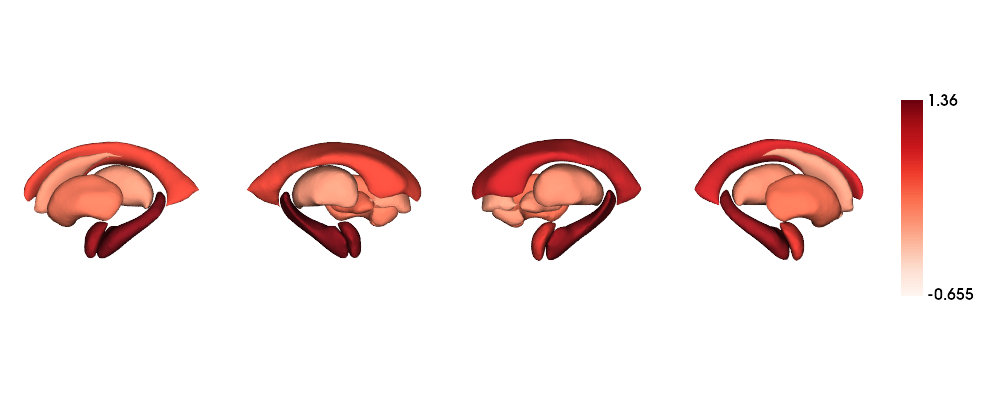}
        \end{minipage}%
        \begin{minipage}{.5\textwidth}
          \centering
          \includegraphics[width=0.49\textwidth,trim={0 4cm 3cm 3cm}]{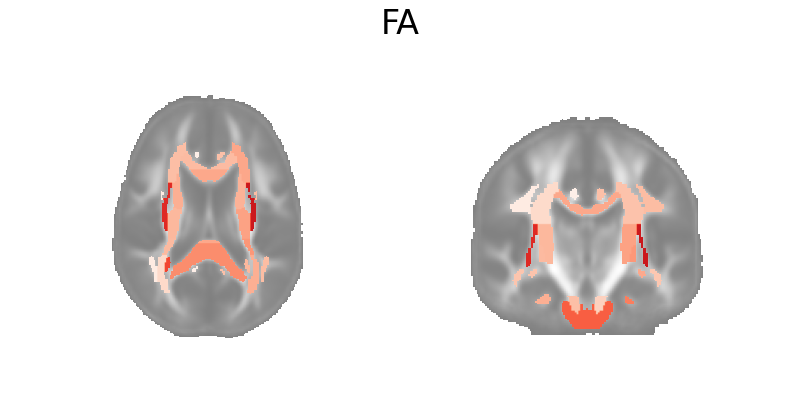}
            \includegraphics[width=0.49\textwidth,trim={0 4cm 3cm 3cm}]{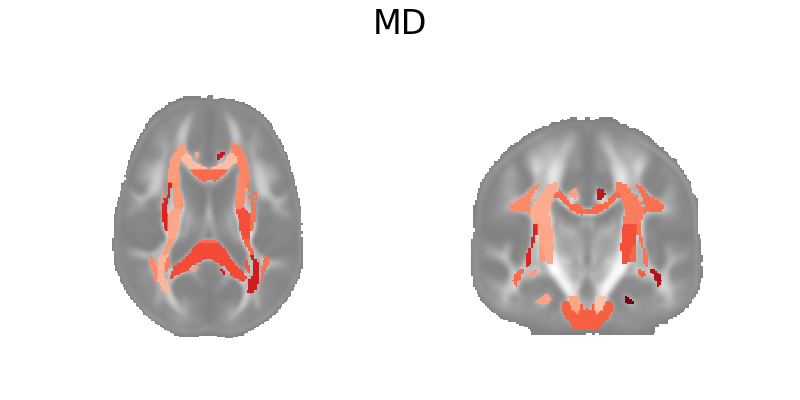}
        \end{minipage}
         \caption{}
          \label{fig:ADNI_AD_recon}
     \end{subfigure}
        \caption{(a) Average $D_{\text{uf}}$ using the gPoE-normVAE ($L_{\text{dim}}$=10) model for the LMCI and (b) AD cohort. The left-hand plots show T1 features and the right-hand plots show DTI features.}\label{fig:deviations_ADNI}
\end{figure}

\textit{Proof.} The probability density of a Gaussian distribution is given by: 

\begin{equation}
p = K \exp\{{-\frac{1}{2}(\textbf{x}-\boldsymbol{\mu})^T\boldsymbol{\Sigma}^{-1}(\textbf{x}-\boldsymbol{\mu})\}} = K\exp{\{\textbf{x}^T\boldsymbol{\Sigma}^{-1}\boldsymbol{\mu}-\frac{1}{2}\textbf{x}^T\boldsymbol{\Sigma}^{-1}\textbf{x}\}}
\end{equation}
where $K$ is a normalisation constant. The product of $N$ Gaussian distributions is a Gaussian of the form: 
\begin{equation}
    \prod_{n=1}^{N} p_n \varpropto \exp{\{\textbf{x}^T \sum_{n=1}^{N} \boldsymbol{\Sigma}^{-1}_{n}\boldsymbol{\mu}_{n}-\frac{1}{2}\textbf{x}^T(\sum_{n=1}^{N}\boldsymbol{\Sigma}^{-1}_{n})\textbf{x}\}}
\end{equation}

where $\boldsymbol{\Sigma}^{-1}\boldsymbol{\mu}=\sum_{n=1}^{N} \boldsymbol{\Sigma}^{-1}_{n}\boldsymbol{\mu}_{n}$ and $\boldsymbol{\Sigma}^{-1} = \sum_{n=1}^{N} \boldsymbol{\Sigma}^{-1}_{n}$. Thus the product Gaussian has a mean $\boldsymbol{\mu} = \frac{\sum_{n=1}^{N}\boldsymbol{\Sigma}^{-1}_{n}\boldsymbol{\mu}_{n}}{\sum_{n=1}^{N} \boldsymbol{\Sigma}^{-1}_{n}}$ and covariance $\boldsymbol{\Sigma}=(\sum_{n=1}^{N} \boldsymbol{\Sigma}^{-1}_{n})^{-1}$. If we have isotropic Gaussian distributions $p_n = \mathcal{N}(\boldsymbol{\mu}_n, \boldsymbol{\sigma}_{n}^{2} \textbf{I})$, the parameters of the product Gaussian become $\boldsymbol{\mu} = \frac{\sum_{n=1}^{N} \boldsymbol{\mu}_{n} / \boldsymbol{\sigma}_{n}^{2}}{\sum_{n=1}^{N} 1 / \boldsymbol{\sigma}_{n}^{2}} \quad \text { and } \quad \boldsymbol{\sigma}^{2} =\frac{1}{\sum_{n=1}^{N} 1 / \boldsymbol{\sigma}_{n}^{2}}$.

Similarly, for a weighted product of $N$ Gaussian distributions, the product Gaussian has the form: 
\begin{equation}
    \prod_{n=1}^{N} p^{\alpha_n}_{n} \varpropto \exp{\{\textbf{x}^T \sum_{n=1}^{N} \alpha_n\boldsymbol{\Sigma}^{-1}_{n}\boldsymbol{\mu}_{n}-\frac{1}{2}\textbf{x}^T(\sum_{n=1}^{N}\alpha_n\boldsymbol{\Sigma}^{-1}_{n})\textbf{x}\}}.
\end{equation}
For isotropic Gaussian distributions, the weighted product Gaussian has mean $
\boldsymbol{\mu} = \frac{\sum_{n=1}^{N} \boldsymbol{\mu}_{n}\boldsymbol{\alpha}_{n} / \boldsymbol{\sigma}_{n}^{2}}{\sum_{n=1}^{N} \boldsymbol{\alpha}_{n} / \boldsymbol{\sigma}_{n}^{2}}$ and variance $ \boldsymbol{\sigma}^{2} = \sum_{n=1}^{N} \frac{1}{ \boldsymbol{\alpha}_{n} / \boldsymbol{\sigma}_{n}^{2}}
$.

\end{document}